\documentclass[]{fairmeta}

\usepackage{amsmath}
\usepackage{amssymb}
\usepackage{amsfonts}
\usepackage{mathtools}
\usepackage{amsthm}
\usepackage{comment}
\usepackage{siunitx}
\usepackage{pifont}
\usepackage{adjustbox}
\usepackage{makecell}
\usepackage{wrapfig}
\usepackage{algorithm}
\usepackage{algorithmicx}
\usepackage{algpseudocode}
\usepackage{float}
\usepackage{listings}
\usepackage{manyfoot}
\usepackage[title]{appendix}
\usepackage{mathrsfs}
\usepackage{colortbl}
\usepackage{tabularx}
\usepackage{enumitem}
\usepackage{dblfloatfix}
\usepackage{capt-of}
\usepackage{nicefrac}
\usepackage{multicol}

\definecolor{tblheader}{RGB}{233,238,245}
\definecolor{tblsubheader}{RGB}{245,247,250}
\definecolor{tblours}{RGB}{238,247,238}
\definecolor{tblreported}{RGB}{244,246,250}
\definecolor{tblna}{gray}{0.55}
\definecolor{tblgrouptext}{RGB}{248,242,232}
\definecolor{tblgrouprl}{RGB}{235,242,249}
\definecolor{tblgroupvisual}{RGB}{243,239,248}
\definecolor{tblgroupmatched}{RGB}{242,244,248}

\usepackage[table]{xcolor}
\definecolor{astracolor}{RGB}{230,120,20}
\def\modelnamebold{\textcolor{astracolor}{\textbf{Astra}}}
\def\modelname{Astra}

\def\modelnamevl{\modelname{}-VL}
\def\modelnamewm{\modelname{}-WM}
\definecolor{navyblue}{HTML}{0071BC}
\newcommand{\up}[1]{{\scriptsize\,(\textcolor{green!45!black}{+#1})}}
\newcommand{\down}[1]{{\scriptsize\,(\textcolor{red!65!black}{-#1})}}
\newcommand{\same}{{\scriptsize\,(\textcolor{gray}{+0.0})}}

\theoremstyle{plain}

\theoremstyle{definition}

\theoremstyle{remark}

\title{%
Thinking with Imagination: Agentic Visual Spatial Reasoning with World Simulators}

\author[1,2*]{Chenming Zhu}
\author[2,3*]{Jingli Lin}
\author[2,4]{Yilin Long}
\author[2,5]{Peizhou Cao}
\author[2]{Tai Wang}
\author[2]{Jiangmiao Pang}
\author[1,\ddagger]{Xihui Liu}

\affiliation[1]{The University of Hong Kong}
\affiliation[2]{Shanghai AI Laboratory}
\affiliation[3]{Shanghai Jiao Tong University}
\affiliation[4]{Fudan University}
\affiliation[5]{Beihang University}

\contribution[*]{Equal contribution}
\contribution[\ddagger]{Corresponding author}

\abstract{While Vision-Language Models (VLMs) have shown strong visual reasoning capabilities, their spatial reasoning abilities remain largely constrained to the observed images and text-oriented chain-of-thought. They often struggle to infer unobserved layouts, maintain cross-view consistency, and reason from alternative viewpoints when only limited egocentric observations are available. In this work, we study this problem as \emph{thinking with imagination}, where a VLM actively acquires imagined visual evidence by interacting with a world simulator during reasoning. We propose \modelnamebold{}, an agentic spatial reasoning framework that empowers VLMs with action-conditioned visual imagination. Specifically, \modelname{} couples \modelname{}-VL, an RL-trained VLM policy, with \modelname{}-WM, a Bagel-based world simulator that generates novel-view observations from context images and natural-language camera motions. To provide reliable imagined evidence, \modelname{}-WM is trained with view consistency tuning to improve pose and content consistency across views. In the RL stage, we propose a world-simulator-in-the-loop two-phase RL curriculum to stabilize tool-use exploration and advance the model's ability to invoke the simulator only when imagined observations improve over direct answering. Experiments demonstrate that both the world simulator and the agentic policy are necessary: \modelname{}-WM improves simulator-augmented Gemini-3-Flash on MMSI-Bench from 45.1 to 49.5, while \modelname{}-VL improves the Qwen3-VL backbone from 29.8 to 38.8 on MMSI-Bench and from 36.8 to 42.7 on MindCube. These results show that imagined observations can provide useful spatial evidence, but effective world-model-augmented reasoning requires learning when, where, and how to imagine.

\par\smallskip\noindent

\begin{tabular}{@{}rl@{}}
    \textbf{Project Page:} \url{https://zcmax.github.io/projects/Thinking-With-Imagination}
\end{tabular}
}
\begin{document}

\maketitle

\section{Introduction}

Spatial reasoning from multi-view images requires more than recognizing visible objects~\cite{mindcube,mmsi,lin2025ostbench,lin2025mmsivideobenchholisticbenchmarkvideobased,viewspatial,3DSRBench,vsibench}. 
When only a few egocentric observations are available, an agent must infer unobserved layout, maintain cross-view consistency, and reason from alternative perspectives. 
A spatial relation may be ambiguous from the current views but become evident after a small viewpoint change, such as moving forward to inspect a camera-object relation, rotating to check viewpoint alignment, or looking beyond the current field of view to infer scene layout. 
Humans naturally handle such uncertainty by maintaining spatial mental models that can be updated and mentally manipulated beyond the observed pixels. 
In contrast, current vision-language models (VLMs) are largely tied to the images they are given, and often struggle to reason about spatial evidence that is missing from the input observations.

Recent studies have explored spatial reasoning scaffolds such as reasoning chains, intermediate views, and cognitive maps~\cite{mindcube,VSI-Bench}.
These approaches suggest that VLMs may benefit from explicit spatial reasoning processes, but most of them still reason over a fixed visual context or rely on pre-defined intermediate representations. 
They do not allow the model to actively decide which missing viewpoint would be useful and acquire corresponding visual evidence during reasoning. 
This motivates an interactive formulation of spatial reasoning, where the model can seek additional evidence instead of only interpreting the observations it already has. 
In this paper, we study this problem as \emph{thinking with imagination}: a VLM issues camera-motion queries to a world simulator, obtains imagined novel-view observations, and continues reasoning over both observed and imagined evidence. 
This transforms spatial mental modeling from a purely internal reconstruction problem into an interactive evidence-acquisition problem.

\begin{figure}[t]
  \centering
  \includegraphics[width=0.95\linewidth]{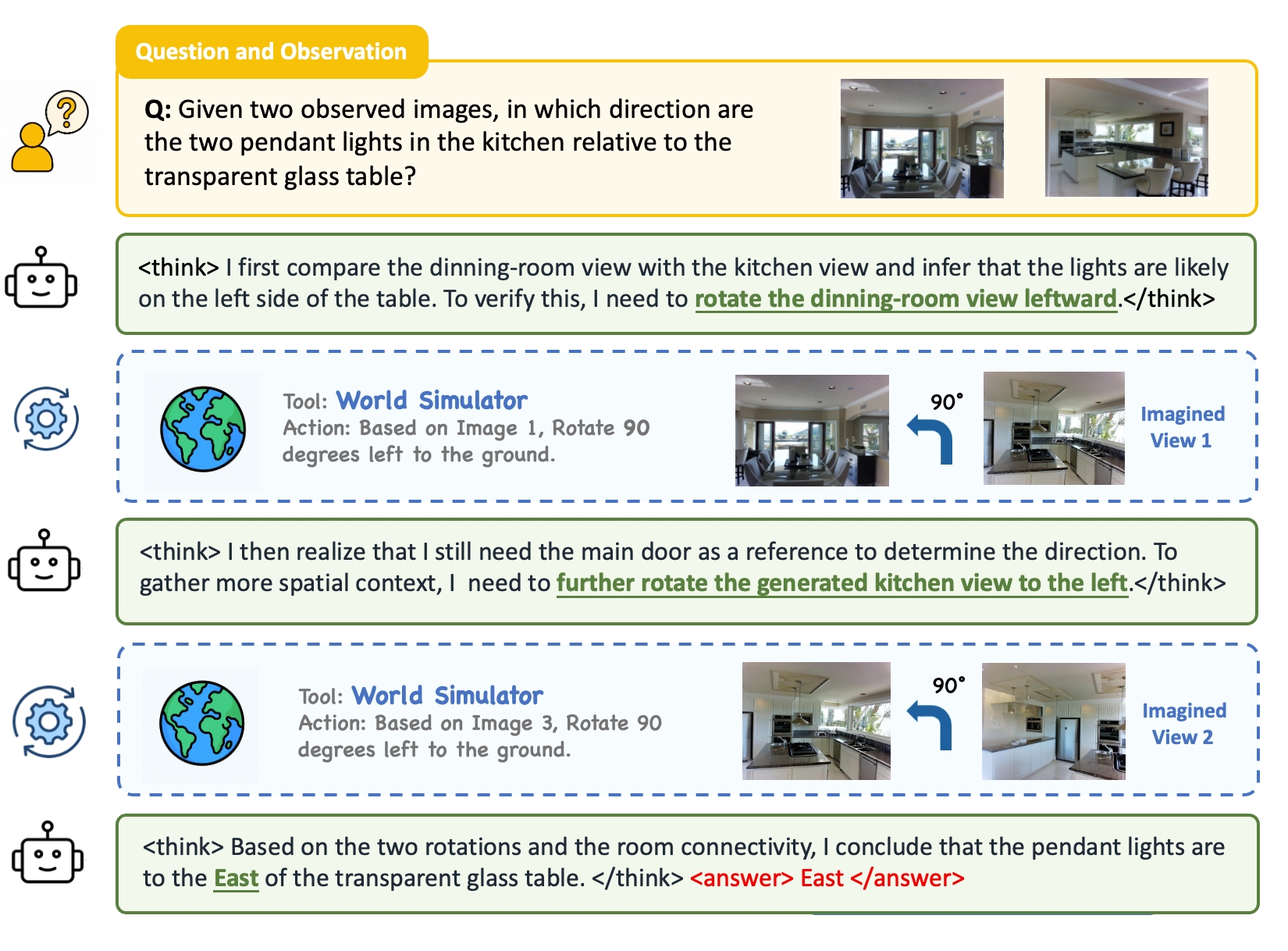}
  \caption{Reasoning trajectory of \modelnamebold{}. \modelname{} tackles the challenging visual spatial reasoning task by agentic leveraging the world simulator within the reasoning process.}
  \label{fig:teaser}
\end{figure}

However, turning this interactive formulation into an effective system cannot be achieved by simply chaining an existing VLM with a generation model. 
First, the world simulator must provide reliable imagined evidence: its outputs should be not only visually plausible, but also spatially consistent with the requested camera motion and the original scene. 
Otherwise, the generated view may mislead the reasoning policy. 
Second, the VLM policy must learn to govern simulator use, including when additional evidence is needed, what viewpoint to request, and how to ground the returned observation in the original context. 
Our preliminary analysis confirms both bottlenecks. 
Off-the-shelf generation models can produce visual plausible images, but often lack spatial consistency to serve as reliable world simulators. 
Meanwhile, naively connecting Qwen3-VL to a world simulator can degrade performance, because the model is not trained to decide when to invoke the simulator, which camera motion to request, or how to incorporate the returned observation.

To address these challenges, we propose \modelnamebold{}, an agentic spatial reasoning framework that integrates an agentic VLM policy with an action-conditioned world simulator. 
The framework consists of two components: \modelname{}-VL, a Qwen3-VL-based agentic reasoning model, and \modelname{}-WM, a Bagel-based world simulator for generating action-conditioned novel views. 
During reasoning, \modelname{}-VL adaptively decides when additional visual evidence is needed, plans a camera-motion query, invokes \modelname{}-WM, and integrates the returned imagined observation into subsequent reasoning. 
This enables \modelname{} to dynamically acquire alternative-view evidence through iterative reasoning and planning, rather than relying only on the initially observed images. We train both components to support reliable agentic imagination. 
For \modelname{}-WM, we construct quality-verified SFT data for world simulation and fine-tune Bagel with View Consistency Tuning, enabling it to generate spatially consistent novel views conditioned on natural-language camera-motion instructions. 
To verify the simulator, we design pose-consistency and content-consistency evaluations, which measure whether generated views follow the requested motion and preserve scene content and spatial layout. 
For \modelname{}-VL, we construct the Spatial QA data corpus and train the policy with a world-simulator-in-the-loop two-phase RL curriculum. 
The first phase teaches valid simulator invocation and keeps tool-use trajectories in the on-policy distribution, while the second phase encourages selective imagination by comparing tool-augmented reasoning with direct no-tool answering.

Our experiments show that world-simulator-augmented spatial reasoning requires both reliable imagined evidence and a policy that can govern its use. 
First, simulator-quality analysis shows that directly using off-the-shelf Bagel brings limited benefit, whereas \modelname{}-WM achieves stronger pose and content consistency and improves simulator-augmented Gemini-3.0-Flash on MMSI-Bench from 45.1 to 49.5. 
This validates the need for spatially consistent world-simulator training rather than generic image generation. 
Second, simulator access alone is not sufficient for open-source VLMs: simply connecting Qwen3-VL to the world simulator can degrade performance, as the model has not learned when to invoke the tool, what camera motion to request, or how to ground the returned observation. 
In contrast, \modelname{}-VL learns world-simulator interaction through RL, and the full \modelname{} framework improves the Qwen3-VL backbone from 29.8 to 38.8 on MMSI-Bench and from 36.8 to 42.7 on MindCube. 
Together, these results show that useful imagination depends on both the quality of the world simulator and the agentic policy that decides when, where, and how to use it. 
Effective imagination is therefore not merely access to a generator, but a learned interaction policy for acquiring and using spatial evidence.

\section{Related Work}

\noindent\textbf{VLMs for Spatial Intelligence.}
Spatial reasoning, which requires understanding and manipulating 3D relationships from visual observations, remains a fundamental challenge for vision-language models (VLMs). Existing efforts mainly improve spatial capability through large-scale training on specialized spatial datasets or by injecting geometric priors into model architectures, such as explicit 3D representations, depth cues, or structure-aware visual features~\cite{sensenova,yang2025visualspatialtuning,zhu2024llava,wang2025ross3d,fan2025vlm3rvisionlanguagemodelsaugmented,hu2025g2vlmgeometrygroundedvision}. While effective, these approaches often rely on static visual inputs and supervised spatial annotations, making it difficult for models to actively resolve ambiguity through interaction. Another line of work equips VLMs with external tools to perform deterministic geometric computation or visual analysis~\cite{chen2025geometricallyconstrainedagentspatialreasoning,han2026tigertoolintegratedgeometricreasoning,zhang2025cooper}. For instance, tool-augmented agents can decompose a spatial query into a sequence of module calls and use the returned results for answer prediction. However, these methods typically treat tools as passive solvers for predefined geometric operations, rather than as interactive environments that can provide new observations conditioned on the model’s own actions. In contrast, our work studies agentic spatial reasoning with the simulator, where the VLM learns to decide when and how to act, query a world simulator for imagined observations, and integrate these observations into multi-turn reasoning to reduce spatial uncertainty.

\noindent\textbf{Thinking with Images.}
Recent works on thinking with images~\cite{o3, deepeyesv2, thinkt_with_images, thyme, openthinkimg} have advanced agentic visual reasoning by enabling VLMs to interleave textual reasoning with iterative visual operations.
Representative systems such as DeepEyes~\cite{deepeyes} show that reinforcement learning can induce image-text reasoning behaviors, while Pixel Reasoner~\cite{pixelreasoner} and Mini-o3~\cite{minio3} reveal that effective tool use often requires staged training to overcome learning traps and discover deep multi-turn trajectories.
SenseNova-MARS~\cite{sensenova} further expands the tool space from perceptual image operations to open-web access and external knowledge fetching.
Beyond operating on existing images, recent studies on multimodal world models suggest that visual generation can serve as an intermediate reasoning representation for physical and spatial inference~\cite{visual_gen}.
While these works demonstrate the promise of image-space tools and generated visuals for reasoning, they primarily study whether visual operations or visual generation can improve reasoning. In contrast, our work formulates world simulation as a general agentic interface for visual spatial reasoning, where a VLM can actively query an action-conditioned world simulator through explicit camera-motion actions and integrate the returned imagined observations into subsequent reasoning.
This formulation decouples the reasoner and planner from the simulator and allows any suitable VLM and world simulator to be composed under a shared action-observation protocol.
It also enables a more detailed study of the interaction between reasoning and simulation, including when to invoke simulation, what actions to request, whether forced simulator use is sufficient, how simulator quality affects downstream performance, and how selective imagination emerges through simulator-in-the-loop reinforcement learning.
\section{Method}
\label{sec:method}

In this section, we describe how \modelnamebold{} couples a spatially consistent world simulator with an agentic VLM policy for interactive spatial reasoning. We first formulate visual spatial reasoning as an interactive decision process in Sec.~\ref{sec:form}. 
We then introduce \modelname{}-WM, a Bagel-based world simulator trained with view consistency tuning to produce spatially consistent imagined observations, in Sec.~\ref{sec:world_simulator}. 
Finally, we present the world-simulator-in-the-loop two-phase RL curriculum for \modelname{}-VL in Sec.~\ref{sec:rl}, followed by the data construction pipeline in Sec.~\ref{sec:data_construct}.

\subsection{Problem Formulation}
\label{sec:form}

\noindent\textbf{Task and Objective.}
Given a spatial question $q$ and an initial set of context images $O_0=\{I_1,\ldots,I_n\}$, the agent must produce the correct answer by reasoning over the available visual evidence. At each turn, it can either answer directly or query a world simulator for an imagined observation from an alternative viewpoint.

\paragraph{Observation Space.}
The interaction trajectory $\mathcal{T}_t$ records all information available to the agent up to turn $t$, including the question, original context images, previous reasoning steps, simulator actions, and simulator outputs. 
We therefore represent the agent state directly as the trajectory:
\[
    s_t = \mathcal{T}_t .
\]

After $t$ simulator calls, the trajectory contains $t$ imagined observations $\{\hat{I}_1,\ldots,\hat{I}_t\}$. 
If the agent invokes the simulator at state $s_t$, the simulator returns a new tool observation
\[
    o_{t+1} = (m_{t+1}, \hat{I}_{t+1}),
\]
where $\hat{I}_{t+1}$ is the generated novel view and $m_{t+1}$ describes the reference image and executed camera motion. 
The trajectory is then updated by appending the new observation:
\[
    \mathcal{T}_{t+1}=\mathcal{T}_t \circ o_{t+1},
\]
where $\circ$ denotes ordered concatenation. 
This motion provenance is important because the policy must distinguish original context images from imagined views and reason about which reference viewpoint and camera motion each generated observation corresponds to.

\noindent\textbf{Action Space.}
At each turn, the agent first produces its reasoning process and then selects one of two high-level actions: 
\textsc{Invoke}, which queries the world simulator for an imagined observation, or \textsc{Answer}, which outputs the final answer and terminates the trajectory. 
Thus, the action space is
\[
    a_t \in \{\textsc{Invoke}(\rho_t), \textsc{Answer}(y_t)\},
\]
where $\rho_t$ denotes the parameters of a simulator query and $y_t$ denotes the final answer.

An \textsc{Invoke} action is parameterized by a reference image, a motion type, and a motion magnitude. 
The motion type is drawn from a compact camera-control vocabulary, including lateral movement, forward/backward movement, yaw rotation, vertical movement, and pitch rotation. 
These parameters are converted into a natural-language camera-motion instruction and sent to the world simulator. 

An \textsc{Answer} action produces the final response and ends the trajectory. 
Trajectories with malformed simulator queries, missing reasoning steps in a single turn are treated as invalid.

\subsection{View Consistency Tuning}
\label{sec:world_simulator}

\modelname{}-WM is the world-simulation component of \modelnamebold{}, serving as the external visual imagination tool for \modelname{}-VL. 
Given the context images, a reference image, and a camera-motion query from the agent, \modelname{}-WM generates a novel-view observation that approximates what the scene would look like after executing the requested motion. 
Formally, when \modelname{}-VL issues an \textsc{Invoke} action with query $\rho_t$, the query specifies a reference image index $r_t$, a motion type, and a motion magnitude. 
We convert $\rho_t$ into a natural-language camera-motion instruction $u_t$ and generate the imagined view as
\[
    \hat{I}_{t+1} = \mathcal{W}(\mathcal{I}_{1:t}, r_t, u_t),
\]
where $\mathcal{I}_{1:t} = \{I_1,\ldots,I_t\}$ denotes the context images available to the agent, $r_t$ indexes the reference image selected by the agent, and $\mathcal{W}$ denotes the Bagel-based \modelname{}-WM.

A useful world simulator must provide spatially consistent evidence rather than merely visually plausible images: the generated view should preserve scene identity, follow the requested camera motion, and maintain the relative layout of visible objects across viewpoints. 
However, off-the-shelf Bagel is not explicitly optimized for this role, and our consistency evaluation shows that its generated views often fail to follow the requested motion or preserve scene content reliably, which can mislead \modelname{}-VL when simulator outputs are treated as visual evidence. 

To address this limitation, we introduce \emph{view consistency tuning}, where Bagel is fine-tuned on quality-verified world-simulator SFT data to generate spatially consistent novel views conditioned on context images and natural-language camera-motion instructions. 
This converts Bagel from a generic multimodal generation model into \modelname{}-WM, an action-conditioned world simulator that better aligns imagined observations with the requested viewpoint change and provides more reliable evidence for downstream spatial reasoning.

\begin{figure}[t]
  \centering
  \includegraphics[width=1\linewidth]{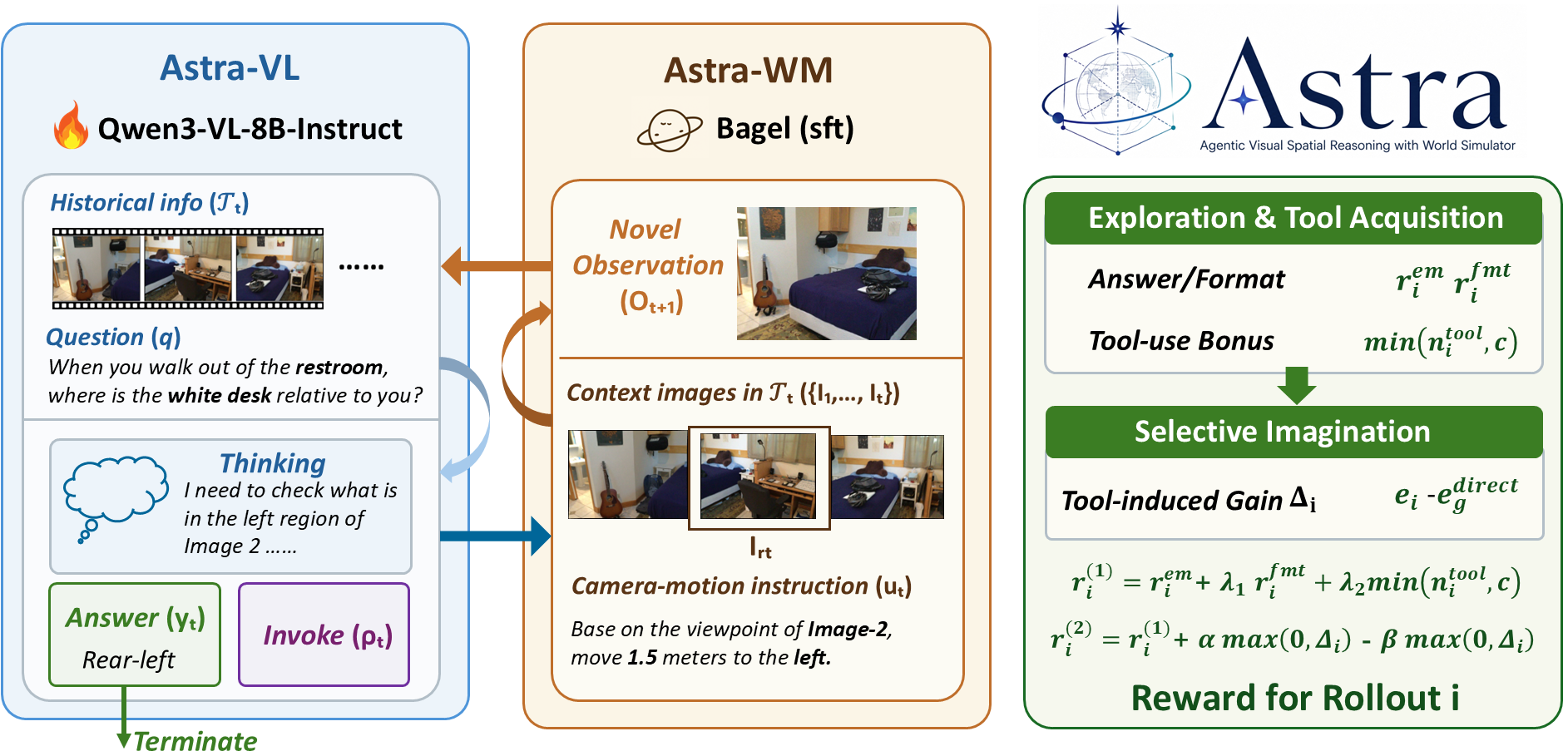}
  \caption{\modelnamebold{} consists of two components: \modelname{}-VL and \modelname{}-WM. The overview illustrates the input-output details of both models during training and inference, as well as the two-phase reinforcement learning training pipeline.}
  \label{fig:overview_astra}
\end{figure}

\subsection{Two-Phase Reinforcement Learning}
\label{sec:rl}

We train \modelname{}-VL with a world-simulator-in-the-loop two-phase RL pipeline: the first phase keeps simulator-invoking trajectories in the on-policy distribution and teaches valid tool interaction, while the second phase encourages the policy to use the simulator only when the imagined observation improves over direct answering. For rollout $i$, let $r^{\mathrm{em}}_i$ denote the final-answer exact-match reward, $r^{\mathrm{fmt}}_i$ denote the format-validity reward, and $n^{\mathrm{tool}}_i$ denote the number of simulator calls in the trajectory. 
Let $\lambda_{\mathrm{fmt}}$ and $\lambda_{\mathrm{use}}$ be the weights for the format reward and simulator-use reward, respectively, and let $c$ denote the maximum number of tool calls eligible for the usage bonus.

\noindent\textbf{Phase 1: Exploration and Tool Acquisition.}
The first phase trains \modelname{}-VL to acquire the mechanics of valid interaction with \modelname{}-WM. 
At this stage, the goal is not yet perfectly selective imagination, but to prevent the policy from collapsing to short direct answers before it has learned how to invoke the simulator. 
We therefore combine the answer reward and format reward with a small capped simulator-use bonus:
\[
r^{(1)}_i =
r^{\mathrm{em}}_i
+ \lambda_{\mathrm{fmt}} r^{\mathrm{fmt}}_i
+ \lambda_{\mathrm{use}} \min(n^{\mathrm{tool}}_i, c).
\]
The cap prevents the model from being rewarded for repeatedly calling the simulator, while still encouraging valid simulator interaction.

\noindent\textbf{Phase 2: Selective Imagination.}
The second phase shifts the objective from invoking the simulator to using it only when it improves spatial reasoning. 
For each rollout group $g$, we additionally evaluate a direct no-tool baseline using a prompt that forbids simulator calls. 
This baseline is used only to estimate the benefit of simulator use and is not included in the policy update. 
For a normal rollout $i$ in group $g$, let $e_i$ be its exact-match score after possible simulator use, and let $e^{\mathrm{direct}}_g$ be the exact-match score of the corresponding no-tool baseline. 
We define the tool-induced gain as
\[
\Delta_i = e_i - e^{\mathrm{direct}}_g .
\]
The phase-two reward combines answer correctness, format validity, optional simulator-use shaping, and the relative gain over direct answering:
\[
r^{(2)}_i =
r^{\mathrm{em}}_i
+ \lambda_{\mathrm{fmt}} r^{\mathrm{fmt}}_i
+ \lambda_{\mathrm{use}} \min(n^{\mathrm{tool}}_i, c)
+ \alpha \max(0,\Delta_i)
- \beta \max(0,-\Delta_i).
\]
Here, $\alpha$ rewards trajectories whose simulator interaction improves over direct answering, while $\beta$ penalizes trajectories whose simulator use hurts the answer.

\subsection{Data Construction}
\label{sec:data_construct}

\noindent\textbf{World Simulator SFT Data.}
Training \modelname{}-WM requires action-conditioned novel-view supervision in the form of context images, a reference image, a camera-motion instruction, and the corresponding target observation.
We construct 544k quality-verified SFT samples from posed multi-view indoor scenes, including IsaacSim~\cite{nvidia_isaac_sim}, ScanNet++~\cite{yeshwanth2023scannet++}, ScanNet~\cite{ScanNet}, Matterport3D~\cite{Matterport3D}, DL3DV~\cite{dl3dv} and ARKitScenes~\cite{arkitscenes}. These samples train \modelname{}-WM to synthesize target views conditioned on scene context and explicit camera-motion queries, matching the interface used by \modelname{}-VL during reasoning. More details are provided in our Appendix.

\noindent\textbf{Agentic RL Data.}
To enhance Qwen3-VL’s ability to invoke tools for improved spatial reasoning, we construct an agentic reinforcement learning dataset from multiple spatial reasoning sources. Specifically, we downsample raw samples from different spatial reasoning categories in SenseNova-800K~\cite{sensenova}, VST-500K~\cite{yang2025visualspatialtuning}, and the training splits of our self-constructed Hard-UMMQA dataset. For each sample, we query Qwen3-VL-8B-Instruct, the model we aim to improve, five times under a high sampling temperature of 1.5. We then retain only the samples for which Qwen3-VL-8B-Instruct produces correct answers in no more than one out of the five trials. These retained examples are regarded as challenging samples and are used for RL training, resulting in 6k  final training samples.
\section{Experiments}
\label{Sec: Experiments}

\subsection{Implementation Details} \label{Implementation_Details}

\noindent\textbf{Model and Training Setup.}
Our \modelnamevl{} is initialized from a Qwen3-VL-8B checkpoint and trained only with the RL stage.
We implement RL training with veRL~\cite{Sheng_2025} and optimize the policy using GRPO. During training, we deploy the \modelnamewm{} through vLLM-Omni~\cite{yin2026vllmomni} as the online world simulator. For RL training, we use a prompt batch size of $128$, a rollout group size of $4$, a PPO mini-batch size of $128$, and a learning rate of $1\times10^{-5}$ for $5$ epochs.
The vision tower is frozen, while the policy is trained with bfloat16 FSDP and gradient checkpointing.
We adopt the Clip-Higher strategy with $\epsilon_{\mathrm{low}}=0.2$ and $\epsilon_{\mathrm{high}}=0.28$, and disable the explicit KL penalty.
Each trajectory supports up to $3$ tool rounds and $10$ assistant turns. We set $\lambda_{\mathrm{fmt}}=0.5$, $\alpha=0.1$, $\beta=0.03$, $c=1$, and $\lambda_{\mathrm{use}}=0.02$. Empirically, this exposes a key tradeoff in agentic spatial RL: sparse gain rewards can cause tool use to collapse, whereas dense usage rewards preserve exploration but may encourage over-imagination.

\noindent\textbf{Evaluation setup.}
We evaluate spatial reasoning on two held-out benchmarks: MMSI-Bench~\cite{mmsi} and MindCube~\cite{mindcube}.
MMSI-Bench contains 1,000 multi-view spatial reasoning examples across diverse spatial-relation categories, while MindCube evaluates multi-view spatial reasoning in structured 3D environments.
We report exact-match answer accuracy (EM) as the main task metric.
For agent behavior, we report the tool-call rate and the average number of tool calls.
For world-simulator quality, we report \textit{pose consistency}, which measures whether generated views follow the requested camera motion, and \textit{content consistency}, which measures object-level precision, recall, and topology consistency to assess scene-content and layout preservation.
More details about benchmark subcategories and simulator-quality metrics are provided in the Appendix.

\vspace{-5pt}

\noindent\textbf{Workflow Modes.}
We evaluate our model against several strong baselines. These include proprietary models, such GLM-4.5V, Gemini-3-Flash, as well as open-souce general VLMs and spatial VLMs. We test the models under the following three workflow settings, which control how a model is allowed to access the world simulator:

\begin{itemize}
\item \textit{Direct Answer}: The model answers directly from the original context images without invoking the world simulator.
\item \textit{Forced Tool-Use}: The model is required by an explicit system prompt to invoke the world simulator for a predefined number of maximum interaction steps.
\item \textit{Agentic Tool-Use}: The model autonomously decides when and how to interact with the world simulator in the rollout reasoning process.
\end{itemize}

\begin{table*}[t]
\caption{
    \textbf{Experimental Results on Spatial Reasoning Benchmarks.}
    We compare \textit{Direct Answer}, \textit{Forced Tool-Use}, and \textit{Agentic Tool-Use} settings.
    Values in parentheses denote absolute changes over the corresponding \textit{Direct Answer} result of the same model.
    \textbf{All} denotes the average accuracy for each benchmark.
    More details on the benchmark subcategories, such as \textbf{PR.}, are provided in the Appendix.
}
\label{tab:main_results_v3}
\begin{center}
\resizebox{\linewidth}{!}{%
\begin{tabular}{c|c|ccccc|cccc}
    \toprule
    \multirow{2}{*}{\textbf{Type}} 
    & \multirow{2}{*}{\textbf{Model}}
    & \multicolumn{5}{c|}{\textbf{MMSI-Bench}}
    & \multicolumn{4}{c}{\textbf{MindCube-tiny}} \\
    \cmidrule(lr){3-7}\cmidrule(lr){8-11}
    & &
    PR. & Attr. & Mot. & MSR & All &
    Rot. & Ard. & Amg. & All \\
    \midrule

    \multicolumn{11}{c}{\textit{\textbf{Direct Answer}}} \\
    \midrule

    \multirow{11}{*}{Open-source}
    & \cellcolor{black!9} Qwen3-VL-8B-Instruct~\cite{Qwen3-VL}
    & \cellcolor{black!9}30.8 & \cellcolor{black!9}30.1 & \cellcolor{black!9}27.7 & \cellcolor{black!9}28.1 & \cellcolor{black!9}29.8
    & \cellcolor{black!9}53.6 & \cellcolor{black!9}38.0 & \cellcolor{black!9}31.1 & \cellcolor{black!9}36.8 \\

    & Qwen3-VL-30B-Instruct~\cite{Qwen3-VL}
    & 31.2 & 35.8 & 25.9 & 29.1 & 30.6
    & 39.9 & 47.5 & 38.5 & 41.8 \\

    & Bagel-7B-MoT~\cite{Bagel}
    & 33.5 & 27.7 & 25.3 & 30.8 & 31.0
    & 34.5 & 31.4 & 42.8 & 34.7 \\

    & SpatialLLM~\cite{SpatialLLM}
    & 24.5 & 23.1 & 22.7 & 30.8 & 25.3
    & 34.0 & 26.8 & 33.0 & 31.1 \\

    & Spatial-MLLM~\cite{Spatialmllm}
    & 28.5 & 25.4 & 18.0 & 26.3 & 26.1
    & 33.8 & 34.5 & 28.3 & 32.1 \\

    & SpatialLadder~\cite{Spatialladder}
    & 30.3 & 23.3 & 16.0 & 21.2 & 25.4
    & 30.5 & 39.8 & 47.8 & 42.3 \\

    & SpaceR~\cite{SpaceR}
    & 29.1 & 29.4 & 21.9 & 22.5 & 26.9
    & 29.8 & 30.0 & 26.8 & 28.3 \\

    & Video-R1~\cite{Video-r1}
    & 30.5 & 25.4 & 22.0 & 26.8 & 27.8
    & 30.0 & 30.5 & 41.3 & 35.8 \\

    & RoboBrain-2.0~\cite{Robobrain2}
    & 28.9 & 28.8 & 22.5 & 28.0 & 28.9
    & 29.7 & 35.8 & 45.2 & 39.6 \\

    & VILASR~\cite{VLASR}
    & 35.9 & 26.0 & 21.0 & 23.2 & 29.8
    & 34.4 & 25.7 & 29.4 & 29.1 \\

    & VLaser~\cite{vlaser}
    & 29.8 & 26.9 & 26.0 & 18.9 & 27.3
    & 31.5 & 24.8 & 38.2 & 32.6 \\

    \midrule

    \multirow{4}{*}{Proprietary}
    & GLM-4.5V~\cite{glm4.5v}
    & 35.6 & 36.9 & 29.3 & 30.3 & 33.8
    & 60.0 & 25.5 & 42.2 & 39.6 \\

    & GPT-4o~\cite{GPT-4o}
    & 28.0 & 32.3 & 36.0 & 30.8 & 30.3
    & 33.5 & 35.0 & 37.2 & 35.8 \\

    & Gemini-2.5-Pro~\cite{Gemini-2.5-and-2.5-Pro}
    & 39.0 & 36.2 & 33.3 & 34.3 & 36.9
    & 89.5 & 54.5 & 48.8 & 57.5 \\

    & Gemini-3-Flash~\cite{gemini3flash}
    & 45.6 & 45.4 & 44.0 & 46.0 & 45.4
    & 93.0 & 72.0 & 61.7 & 70.5 \\

    \midrule
    \multicolumn{11}{c}{\textit{\textbf{Forced Tool-Use (zero-shot)}}} \\
    \midrule

    \multirow{3}{*}{Open-source}
    & \cellcolor{black!9} Qwen3-VL-8B-Instruct~\cite{Qwen3-VL}
    & \cellcolor{black!9}30.4\down{0.4} & \cellcolor{black!9}29.5\down{0.6} & \cellcolor{black!9}19.6\down{8.1} & \cellcolor{black!9}30.8\up{2.7} & \cellcolor{black!9}28.6\down{1.2}
    & \cellcolor{black!9}31.1\down{22.5} & \cellcolor{black!9}23.7\down{14.3} & \cellcolor{black!9}26.8\down{4.3} & \cellcolor{black!9}27.6\down{9.2} \\

    & Qwen3-VL-30B-Instruct~\cite{Qwen3-VL}
    & 31.5\up{0.3} & 28.7\down{7.1} & 21.6\down{4.3} & 28.1\down{1.0} & 28.9\down{1.7}
    & 34.7\down{5.2} & 32.7\down{14.8} & 38.1\down{0.4} & 35.7\down{6.1} \\

    & Bagel-7B-MoT~\cite{Bagel}
    & 31.3\down{2.2} & 25.6\down{2.1} & 24.7\down{0.6} & 28.7\down{2.1} & 29.7\down{1.3}
    & 33.9\down{0.6} & 26.8\down{4.6} & 31.8\down{11.0} & 29.2\down{5.5} \\

    \midrule

    \multirow{1}{*}{Proprietary}
    & Gemini-3-Flash~\cite{gemini3flash}
    & 50.4\up{4.8} & 51.5\up{6.1} & 43.4\down{0.6} & 50.3\up{4.3} & 49.5\up{4.1}
    & 93.0\same & 70.3\down{1.7} & 65.0\up{3.3} & 72.7\up{2.2} \\

    \midrule
    \multicolumn{11}{c}{\textit{\textbf{Agentic Tool-Use}}} \\
    \midrule

    \multirow{1}{*}{Open-source}
    & \cellcolor{navyblue!9} \modelname{} \textsubscript{Qwen3-VL-8B-Instruct}
    & \cellcolor{navyblue!9}\textbf{42.3}\up{11.5}
    & \cellcolor{navyblue!9}\textbf{41.0}\up{10.9}
    & \cellcolor{navyblue!9}\textbf{32.1}\up{4.4}
    & \cellcolor{navyblue!9}\textbf{33.6}\up{5.5}
    & \cellcolor{navyblue!9}\textbf{38.8}\up{9.0}
    & \cellcolor{navyblue!9}\textbf{60.1}\up{6.5}
    & \cellcolor{navyblue!9}\textbf{43.5}\up{5.5}
    & \cellcolor{navyblue!9}\textbf{36.8}\up{5.7}
    & \cellcolor{navyblue!9}\textbf{42.7}\up{5.9} \\

    \bottomrule
\end{tabular}
}
\end{center}
\vspace{-1em}
\end{table*}

\subsection{Main Results}
\label{sec:results}

\noindent\textbf{Effectiveness of \modelnamewm{}.}
We first evaluate whether generated observations from the world simulator provide useful evidence for spatial reasoning under the \textit{Forced Tool-Use} setting.
In this setting, the reasoning model is required to invoke the world simulator for a predefined number of steps before producing the final answer. We evaluate Gemini-3-Flash with different world simulators under the forced two-step tool-use setting.
As shown in Tab.~\ref{tab:gemini_bagel}, off-the-shelf Bagel provides only limited gains, whereas our trained \modelnamewm{} substantially improves simulator-augmented Gemini-3-Flash on MMSI-Bench, increasing the overall accuracy from 0.451 to 0.495 and the spatial-relation accuracy from 0.458 to 0.504.
This shows that imagined observations can provide useful evidence for downstream spatial reasoning, but their benefit depends on the spatial consistency of the simulator. However, a reliable simulator alone is not sufficient. Prompted forced tool-use requires the model to interact with the simulator mechanically, but does not teach it whether additional observations are needed, which viewpoint would be most informative, or how the returned observation should be grounded in the original context.
As shown in Tab.~\ref{tab:main_results_v3}, open-source VLMs such as Qwen3-VL and Bagel can underperform their direct-answer counterparts under forced tool-use, indicating that they often fail to generate useful simulator actions or integrate imagined observations effectively.

\begin{table*}[t]
\caption{
    \textbf{Simulator Quality and Zero-Shot Simulator-Augmented Reasoning on MMSI-Bench.} 
}
\label{tab:gemini_bagel}
\begin{center}
\resizebox{\linewidth}{!}{%
\begin{tabular}{lcc|cccccccc}
\toprule
\textbf{Model} 
& \textbf{Pose Cons.} 
& \textbf{Content Cons.} 
& \textbf{All} 
& \textbf{PR.} 
& \textbf{Cam.--Obj.} 
& \textbf{Cam.--Reg.} 
& \textbf{Cam.--Cam.} 
& \textbf{Obj.--Obj.} 
& \textbf{Obj.--Reg.} 
& \textbf{Reg.--Reg.} \\ 
\midrule

\rowcolor{black!6}
Gemini-3-Flash 
& -- 
& -- 
& 0.451 
& 0.458 
& 0.442 
& \textbf{0.602} 
& 0.430 
& 0.436 
& 0.447 
& 0.395 \\

+ Bagel
& 9.0/3.0 
& 0.356/0.396/0.102 
& 0.458 
& 0.469 
& \textbf{0.523} 
& 0.434 
& 0.443 
& 0.483 
& 0.459 
& 0.468 \\

+ Astra-WM$_{\text{30k}}$
& \textbf{72.5}/70.5 
& 0.532/0.560/0.230 
& 0.463 
& 0.471 
& 0.510 
& 0.470 
& 0.411 
& 0.511 
& 0.458 
& 0.470 \\

\rowcolor{navyblue!9}
+ Astra-WM$_{\text{60k}}$
& 69.0/\textbf{75.0} 
& \textbf{0.534/0.561/0.234} 
& \textbf{0.495} 
& \textbf{0.504} 
& 0.500 
& 0.554 
& \textbf{0.554} 
& \textbf{0.522} 
& \textbf{0.470} 
& \textbf{0.494} \\

\bottomrule
\end{tabular}
}
\end{center}
\vspace{-1em}
\end{table*}

\noindent\textbf{Effectiveness of \modelnamevl{}.}
We next study whether an open-source VLM can be trained to use the world simulator in an agentic manner.
Unlike forced tool-use, which prescribes a fixed number of simulator interactions, \modelnamebold{} makes tool-use decisions conditioned on the current reasoning state.
This requires the policy to solve three coupled problems: deciding whether additional evidence is needed, selecting an informative camera-motion query, and grounding the returned imagined observation in the original context before answering.
As shown in Tab.~\ref{tab:main_results_v3}, simply prompting Qwen3-VL to use the simulator does not reliably improve performance and can even underperform direct answering, suggesting that simulator access alone is insufficient for open-source VLMs.
In contrast, after world-simulator-in-the-loop RL training, \modelnamevl{} learns to interact with the simulator effectively, and the full \modelnamebold{} framework improves the Qwen3-VL-8B backbone from 29.8 to 38.8 on MMSI-Bench and from 36.8 to 42.7 on MindCube.
These gains indicate that the improvement comes not from tool access alone, but from learning an agentic policy that can decide when to imagine, where to query, and how to use imagined evidence.

\subsection{Ablation Study}
\label{sec:ablation}

We ablate three factors that determine the effectiveness of \modelnamebold{}:
the spatial consistency of the world simulator, the reward design for maintaining useful tool exploration, and the inference-time control over simulator interaction.
Together, these ablations show that effective imagination requires more than simulator access alone: the imagined observations must be spatially reliable, and the policy must learn when and how to use them.

\noindent\textbf{Effectiveness of View Consistency Tuning.}
The goal of training the world simulator is not merely photorealistic generation, but spatially reliable evidence for reasoning.
As shown in Tab.~\ref{tab:gemini_bagel}, off-the-shelf Bagel has low pose and content consistency, and therefore brings only limited and uneven gains when used as a zero-shot simulator.
After view consistency tuning, \modelname{}-WM produces more reliable imagined observations, with substantially improved motion following and scene-content preservation.
This improvement in simulator quality leads to stronger downstream spatial reasoning, yielding the best overall and spatial-relation performance among the simulator variants.
The gains are especially evident on relation types that benefit from alternative viewpoints, such as Cam.--Cam., Obj.--Obj., Obj.--Reg., and Reg.--Reg.
These results show that generic image generation is insufficient for world-simulator-augmented reasoning; useful imagined observations require both accurate camera-motion following and consistent scene-layout preservation.

\begin{table*}[t]
\caption{
\textbf{Effectiveness of the Two-Phase RL Curriculum on MMSI-Bench.}
}
\label{tab:two_phase_rl_ablation}
\centering
\resizebox{\linewidth}{!}{%
\begin{tabular}{l|c|cc|ccccc}
\toprule
\multirow{2}{*}{\textbf{Type}} 
& \multirow{2}{*}{\textbf{Method}}
& \multicolumn{2}{c|}{\textbf{Tool Behavior}} 
& \multicolumn{5}{c}{\textbf{MMSI-Bench}} \\
\cmidrule(lr){3-4}\cmidrule(lr){5-9}
&
& \textbf{Tool rate} & \textbf{Calls/row} 
& \textbf{PR.} & \textbf{Attr.} & \textbf{Mot.} & \textbf{MSR} & \textbf{All} \\
\midrule

\multirow{2}{*}{\textit{Single-stage}}
& Tool-gain only ($\lambda_{\mathrm{use}}=0$)
& 4.9 & 0.049
& 36.5 & 36.8 & 28.4 & 31.2 & 34.3 \\

& Usage bonus only ($\lambda_{\mathrm{use}}=0.02$)
& 98.1 & 1.400
& 39.4 & 38.7 & 30.2 & 32.5 & 36.1 \\

\midrule

\multirow{2}{*}{\textit{Two-phase}}
& Phase 1 only
& 98.0 & 1.120
& 40.1 & 39.2 & 30.8 & 32.9 & 36.8 \\

& \cellcolor{navyblue!9}\textbf{Phase 1 $\rightarrow$ Phase 2 (full)}
& \cellcolor{navyblue!9}\textbf{61.5} 
& \cellcolor{navyblue!9}\textbf{0.780}
& \cellcolor{navyblue!9}\textbf{42.3} 
& \cellcolor{navyblue!9}\textbf{41.0} 
& \cellcolor{navyblue!9}\textbf{32.1} 
& \cellcolor{navyblue!9}\textbf{33.6} 
& \cellcolor{navyblue!9}\textbf{38.8} \\

\bottomrule
\end{tabular}
}
\vspace{-1em}
\end{table*}

\noindent\textbf{Effectiveness of the Two-Phase RL Curriculum.}
Tab.~\ref{tab:two_phase_rl_ablation} studies how reward design affects both tool-use behavior and downstream spatial reasoning.
A single-stage tool-gain reward provides a direct comparison against no-tool answering, but it fails to maintain exploration.
The policy quickly collapses toward direct answering, reaching only a 4.9\% tool-call rate and 0.049 calls per rollout.
This confirms that sparse relative-gain rewards are too weak early in training, when valid and useful simulator trajectories are still rare. Adding a dense simulator-use bonus prevents this collapse, but leads to the opposite failure mode.
The policy invokes the simulator on 98.1\% of rollouts with 1.400 calls per rollout, indicating near-universal imagination rather than selective tool use.
Although this improves over the collapsed setting, it does not teach the model when the simulator is actually necessary.
Similarly, the Phase-1-only model learns the mechanics of simulator interaction, but still relies on frequent tool use and lacks a reliable stopping criterion. Our full two-phase curriculum achieves the best balance between exploration and selective imagination.
It obtains the strongest MMSI-Bench performance, improving the overall score to 38.8 and achieving the best results.
These results show that the two phases play complementary roles:
Phase 1 keeps simulator-invoking trajectories within the on-policy distribution and teaches valid tool interaction, while Phase 2 shifts the objective from merely calling the simulator to using it only when imagined observations improve over direct answering.
Thus, the gain of \modelname{} does not come from encouraging more tool calls, but from learning a more effective policy for deciding when simulator interaction is useful.

\begin{figure}[t]
  \centering
  \includegraphics[width=1\linewidth]{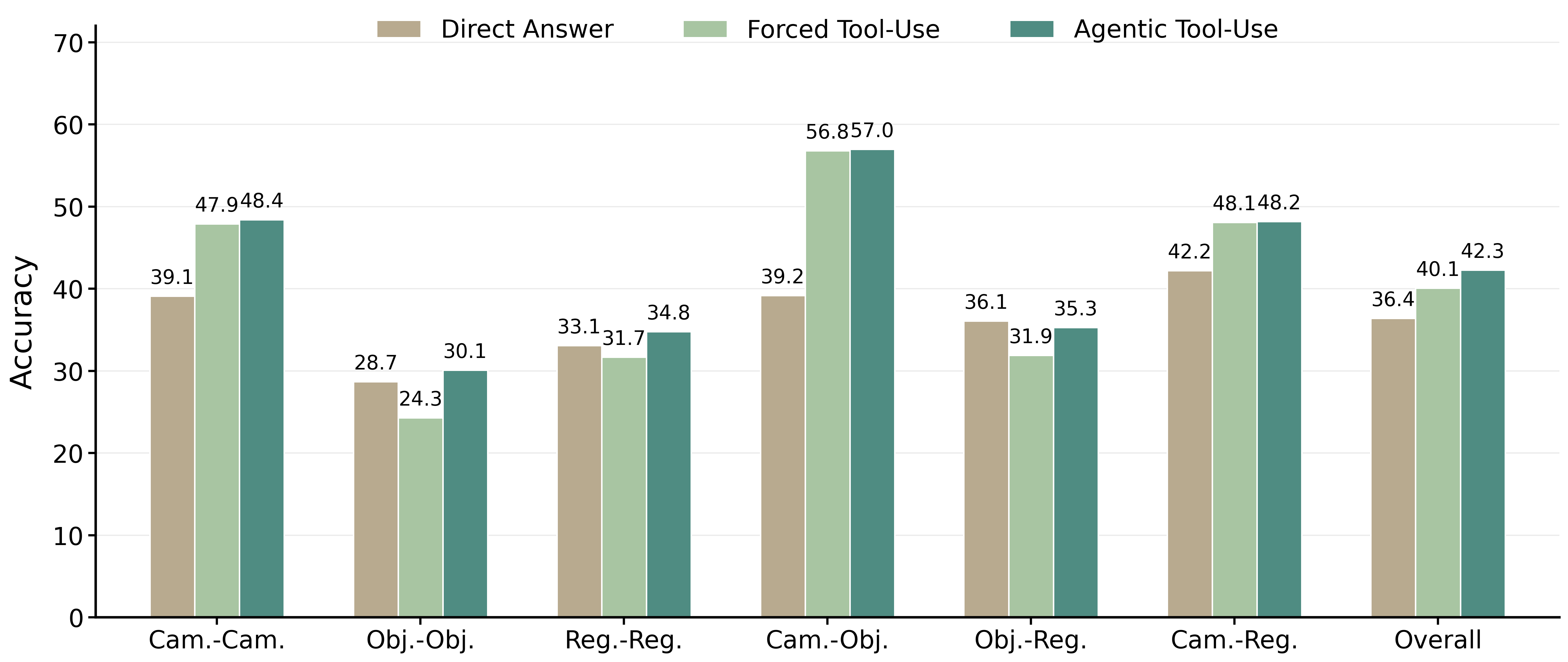}
  \caption{
  \textbf{Inference-time workflow mode ablation of our \modelname{} on MMSI-Bench.}
  }
  \label{fig:interaction_mode_ablation}
\end{figure}

\noindent\textbf{Inference-Time Workflow Mode Ablation.}
We further ablate the inference-time workflow mode of the same trained \modelname{} policy on MMSI-Bench positional-relation tasks.
This experiment isolates the effect of agentic control at inference time:
the model parameters and world simulator are fixed, while only the workflow mode is changed among \textit{Direct Answer}, \textit{Forced Tool-Use}, and \textit{Agentic Tool-Use}, as defined above. As shown in Fig.~\ref{fig:interaction_mode_ablation}, simulator access affects different relation types unevenly.
Forced tool-use substantially improves camera-centric relations, increasing Cam.--Cam. from 39.1 to 47.9, Cam.--Obj. from 39.2 to 56.8, and Cam.--Reg. from 42.2 to 48.1.
This suggests that imagined viewpoints are especially useful when the question depends on camera pose, egocentric viewpoint changes, or missing visual evidence from an alternative view.
However, forced tool-use hurts several object- or region-centric relations:
Obj.--Obj. drops from 28.7 to 24.3, Reg.--Reg. drops from 33.1 to 31.7, and Obj.--Reg. drops from 36.1 to 31.9.
These categories often require stable reasoning over the existing layout, where unnecessary generated views may introduce noise or distract the policy from reliable original observations. Agentic tool-use achieves the best overall performance, improving from 36.4 in direct-answer mode and 40.1 in forced tool-use mode to 42.3.
It preserves the large gains on camera-centric categories while recovering performance on object- and region-centric categories compared with forced tool-use. This shows that the benefit of \modelname{} comes not only from access to imagined observations, but also from the learned policy's ability to decide whether to invoke the simulator, which action to issue, and how to ground the final answer in the returned observation.

\section{Conclusion}
\label{Sec: Conclusions}

We introduced \modelname{}, an agentic spatial reasoning framework that couples a spatially consistent world simulator with an RL-trained VLM policy. 
By querying imagined observations from alternative viewpoints, \modelname{} enables VLMs to acquire missing spatial evidence rather than reasoning only from the given images. 
Our experiments show that both reliable simulation and selective tool-use policy learning are necessary: \modelname{}-WM improves the usefulness of generated views through view consistency tuning, while \modelname{}-VL learns when and how to invoke the simulator through a two-phase RL curriculum. 
These results highlight that effective imagination is not simply access to a generator, but a learned interaction process for acquiring, grounding, and using spatial evidence.

\bibliographystyle{unsrtnat}
\bibliography{main}

\beginappendix
\newpage
\appendix

\section{Additional Details on Training Data}

\subsection{World Simulator SFT Data}

\begin{figure}[t]
  \centering
  \includegraphics[width=1\linewidth]{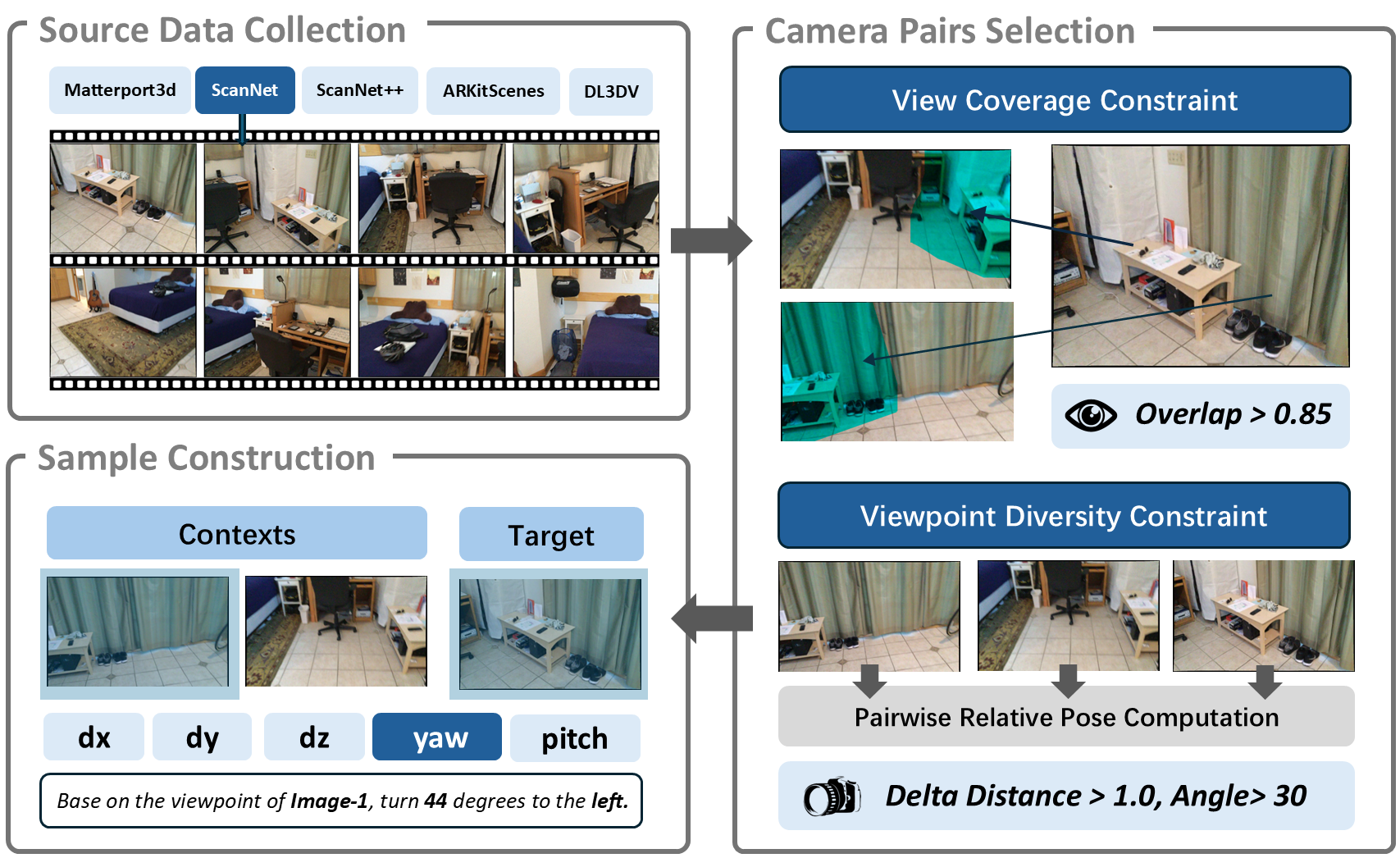}
  \caption{Pipeline for Constructing Training Data for the World Simulator}
  \label{fig:world_model_sft_data_pipeline}
\end{figure}

To equip the world simulator (i.e., Bagel) with strong novel view synthesis capabilities, we construct a large-scale training dataset consisting of tuples $(\mathcal{I}_{ctx}, \; p, \; I_{tgt})$, where $\mathcal{I}_{ctx}$ denotes a set of context images, $p$ is a camera motion instruction (e.g., ``move 2.5 meters to the left''), and $I_{tgt}$ is the target image corresponding to the new viewpoint. Given $\mathcal{I}_{ctx}$ and $p$, the model is required to infer the scene geometry and synthesize the observation $I_{tgt}$ under the transformed camera pose. An overview is illustrated in Fig.~\ref{fig:world_model_sft_data_pipeline}:

\textbf{Source Data Collection}. Our training scenes are collected from ScanNet, ScanNet++, Matterport3D, ARKitScenes, and DL3DV (training splits), covering both indoor and outdoor environments, with a total of 11,292 scenes. Each scene is represented as a scanned RGB-D video, consisting of per-camera tuples $\{(I_i, D_i, \mathbf{T}_i)\}_{i=1}^{M}$, where $I_i$ is the RGB image, $D_i$ is the depth image, and $\mathbf{T}_i \in SE(3)$ is the camera pose.

\textbf{Camera Pairs Selection}. For each scene, we sample camera pairs to construct training samples $(\mathcal{C}_{ctx}, \; C_{tgt})$,
where $\mathcal{C}_{ctx}$ contains 2--3 context cameras and $C_{tgt}$ is the target camera. To ensure the validity and effectiveness of the constructed samples, each camera pair must satisfy the following two constraints:
\begin{itemize}
    \item \textit{View Coverage Constraint.} To guarantee the completeness of background information in the target view, 
we require that most of the visible scene in the target camera can be observed from the context cameras. The visible point cloud of a camera is obtained via back-projection:
\begin{equation}
\mathcal{P}(C) = \left\{ \mathbf{X} = \mathbf{T} \cdot \left( D(u,v) \cdot K^{-1}[u, v, 1]^T \right) \right\}.
\end{equation}
The coverage ratio is defined as:
\begin{equation}
\text{Coverage} = 
\frac{
|\mathcal{P}(C_{tgt}) \cap \bigcup\limits_{C \in \mathcal{C}_{ctx}} \mathcal{P}(C)|
}{
|\mathcal{P}(C_{tgt})|
}.
\end{equation}

We require $\text{Coverage} \geq 0.85$.

\item \textit{Viewpoint Diversity Constraint.} To avoid redundancy in visual information, we require sufficient differences between cameras in the same pair. For any pair of cameras $C_i, C_j \in \mathcal{C}_{ctx}$, at least one of the following conditions must hold:
\begin{equation}
\|\mathbf{t}_i - \mathbf{t}_j\|_{\text{xy}} \geq 1.0 \;\text{m}
\quad \text{or} \quad
|z_i - z_j| \geq 1.0 \;\text{m},
\end{equation}
\begin{equation}
|\Delta \theta_{ij}| \geq 30^\circ
\quad \text{or} \quad
|\Delta \phi_{ij}| \geq 30^\circ,
\end{equation}
where $\Delta \theta_{ij}$ and $\Delta \phi_{ij}$ denote the yaw (left-right rotation) and pitch (up-down rotation) differences, respectively.

\end{itemize}

\textbf{Sample Construction. } Given a camera pair $(\mathcal{C}_{ctx}, C_{tgt})$, 
we select one context camera as the source camera $C_{src}$. We compute the relative camera transformation:
\begin{equation}
\Delta \mathbf{T} = \mathbf{T}_{tgt} \mathbf{T}_{src}^{-1},
\end{equation}
and decompose it into interpretable motion components $(d_x, d_y, d_z, d\theta, d\phi)$, where $(d_x, d_y)$ denote horizontal-plane translation, $d_z$ denotes vertical motion, and $(d\theta, d\phi)$ denote yaw and pitch rotations. These motion parameters are converted into natural language prompts $p$. (To simplify the problem into a one-step setting, we filter the data and retain only samples where a single degree of freedom dominates, i.e., only one of the five motion components changes significantly). Based on the above pipeline, we construct a total of 544,197 training samples.

\section{World Simulator Evaluation}
To evaluate the world simulator's novel-view generation capability, we construct a test set using the same pipeline as for the World Simulator SFT data. Specifically, we sample from the test splits of five datasets, including DL3DV~\cite{dl3dv}, ScanNet~\cite{ScanNet}, ScanNet++~\cite{yeshwanth2023scannet++}, Matterport3D~\cite{Matterport3D}, and ARKitScenes~\cite{arkitscenes}. For each dataset, we collect 200 samples, resulting in 1,000 evaluation samples in total. We evaluate the generated views from two complementary perspectives: pose consistency and content consistency.

\noindent\textbf{Pose Consistency.}
The pose metric evaluates whether the generated image reflects the intended camera motion. Since the world simulator only produces an RGB image, we first estimate the depth of the generated image using Depth Anything, and then recover its corresponding camera pose by aligning it with the source RGB-D observation and the known source camera pose. Based on the predicted target pose and the ground-truth target pose, we compute their relative transformations with respect to the source camera.

Following the motion representation used in World Simulator SFT data construction, each relative transformation is decomposed into five interpretable components:
\[
(d_x, d_y, d_z, d_{\theta}, d_{\phi}),
\]
where $(d_x,d_y,d_z)$ denote translational changes and $(d_{\theta},d_{\phi})$ denote yaw and pitch rotations. We compare the predicted motion with the ground-truth motion along each degree of freedom. A generated sample is considered pose-consistent if it correctly preserves unchanged dimensions and predicts the correct direction and a sufficiently close magnitude for the dominant changed dimension. The final pose score is computed as the average success rate over all evaluation samples.

\noindent\textbf{Content Consistency.}
The content metric evaluates whether the generated image preserves the visual information of the ground-truth target image, including object categories, object counts, object locations, and pairwise spatial relations. We first use a VLM to identify the union of key object categories appearing in either the generated image or the target image, using the following prompt:

\begin{quote}
\small
\textit{Please list ALL distinct key object categories that appear in EITHER image. Use exactly the same category name for the same type of object across both images.}
\end{quote}

Given the extracted object categories, we then use GroundingDINO~\cite{liu2024groundingdinomarryingdino} to detect the corresponding objects in both images. Detected objects are matched between the generated image and the target image according to category consistency and bounding-box overlap. Based on the matched objects, we compute object-level recall and precision, which respectively measure how many target objects are successfully generated and how many generated objects are consistent with the target image.

In addition to object-level consistency, we further evaluate spatial topology consistency. For each pair of objects in the target image, we represent their spatial relation using the direction between their bounding-box centers. If both objects can be matched in the generated image, we compare the corresponding pairwise relation with that in the target image; otherwise, the relation is treated as missing. The topology score is computed by averaging this consistency over all object pairs.

Together, the pose and content metrics provide a holistic evaluation of whether the world simulator can follow the specified camera motion while generating a target view with consistent object-level content and spatial structure.

\section{Case Study and Error Analysis}
\label{app:case_study_error_analysis}

We provide qualitative case studies to better understand how the visual-CoT trajectory succeeds or fails when interacting with the world simulator.
Rather than assigning each failure to a single category, we analyze representative cases along the full reasoning chain: the initial spatial hypothesis, the tool-use decision, the generated action, the simulator observation, and the final answer.
These cases further illustrate why effective world-model-augmented reasoning requires more than tool access.

\begin{figure}[t]
  \centering
  \includegraphics[width=0.8\linewidth]{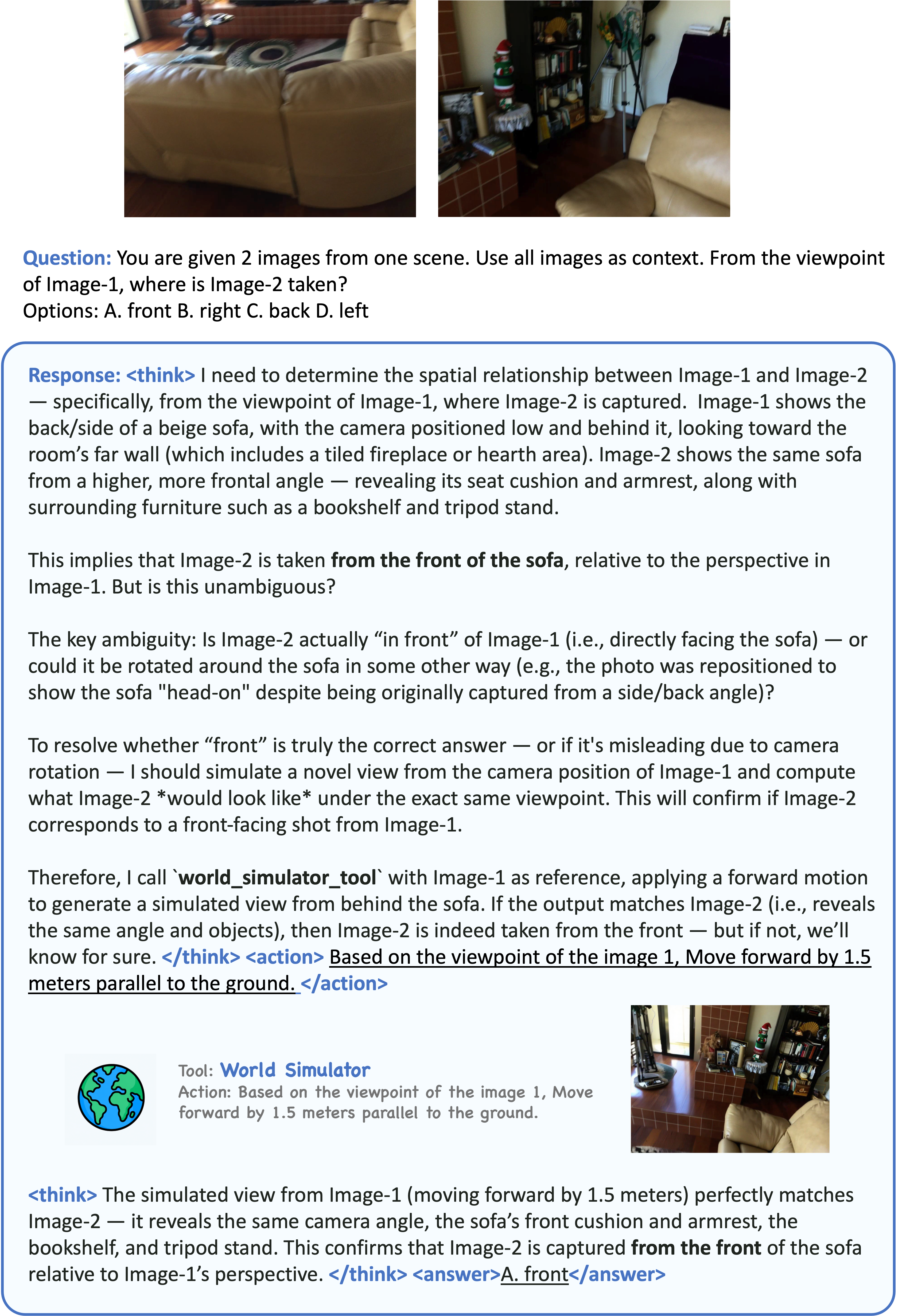}
  \caption{Case 1 of \modelnamebold{}.}
  \label{fig:case1}
\end{figure}

\noindent\textbf{Case 1: Imagination Provides Missing Viewpoint Evidence.}
In successful cases, the original context images leave an important spatial relation ambiguous, but the model identifies the uncertainty and queries a viewpoint that directly resolves it.
The returned observation provides new visual evidence that was not available in the original context, allowing the model to \textit{revise or confirm its spatial hypothesis before answering}, as illustrated in Fig.~\ref{fig:case1}.
This case shows the intended behavior of agentic imagination: the tool is not used as a generic extra image generator, but as a targeted evidence-acquisition mechanism.

\noindent\textbf{Case 2: Correct Tool Access But Uninformative Action.}
In some failures, the model correctly decides that additional evidence is needed, but issues an action that does not reduce the relevant uncertainty.
For example, it may rotate in a direction that keeps the target object out of view, move from an unhelpful reference image, or request a viewpoint change that is unrelated to the queried relation.
The resulting observation may be spatially consistent, but it does not help answer the question.
This suggests that the bottleneck is not only whether to call the simulator, but whether the model can formulate informative camera-motion queries.

\begin{figure}[t]
  \centering
  \includegraphics[width=0.8\linewidth]{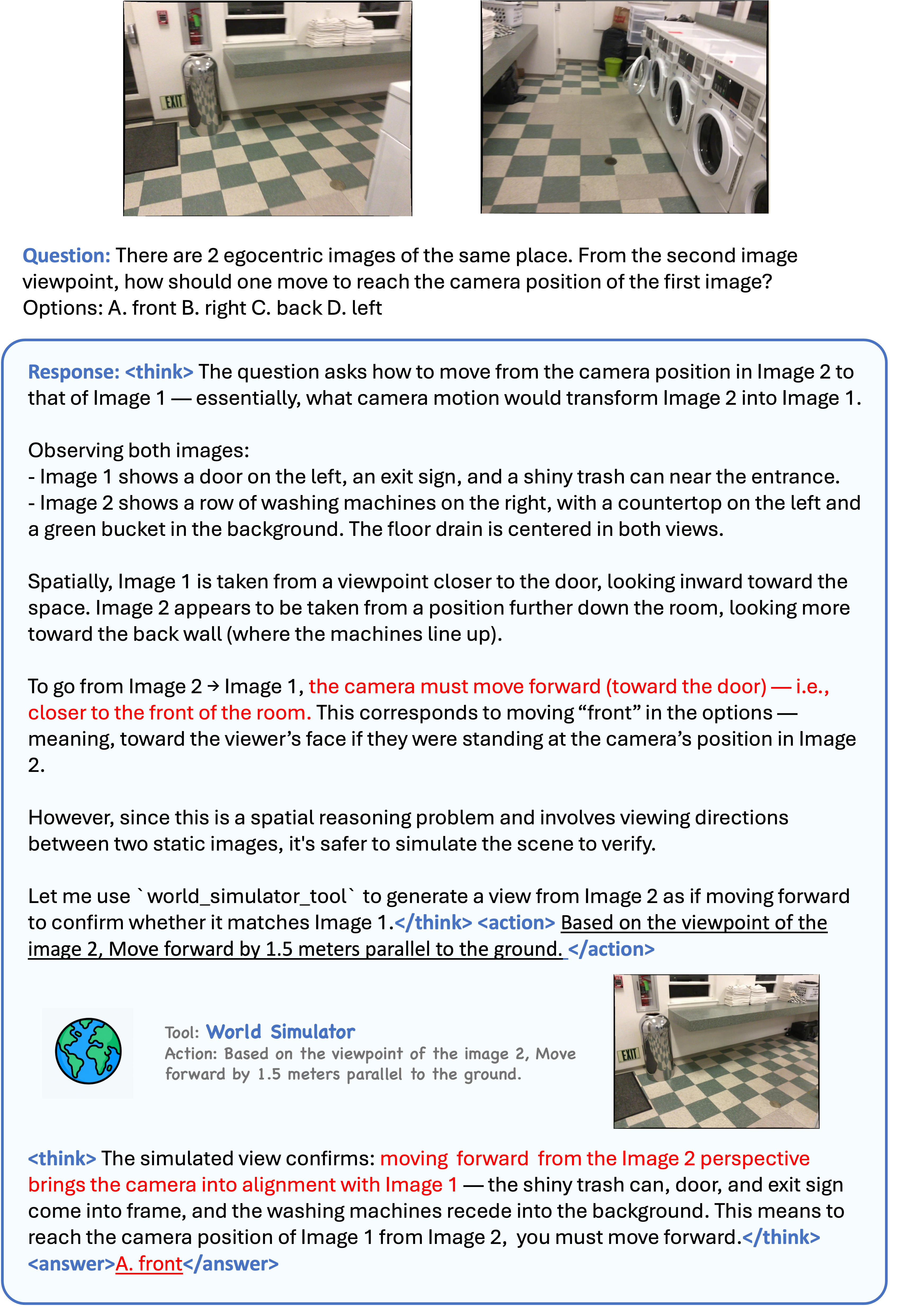}
  \caption{Case 2 of \modelnamebold{}.}
  \label{fig:case2}
\end{figure}

\noindent\textbf{Case 3: Spatially Inconsistent Simulator Observation.}
Some failures originate from the world simulator itself.
The model may request a camera motion, but the generated observation does not faithfully follow the instruction or fails to preserve the scene layout.
Typical issues include incorrect viewpoint change, missing or hallucinated objects, object identity drift, or layout inconsistency across views.
Because the reasoning policy treats the generated image as evidence, such simulator errors can mislead the final answer. As illustrated in Fig.~\ref{fig:case2}, the model makes an incorrect spatial analysis during its initial attempt to reason about the spatial relation. The subsequently obtained simulated observation is spatially inconsistent, which further reinforces the model's earlier mistaken spatial interpretation rather than correcting it.

\noindent\textbf{Case 4: Useful observation is ignored or misused.}
In other cases, the simulator returns a potentially useful observation, but the model fails to incorporate it correctly.
The model may continue relying on its initial hypothesis, confuse original and generated image indices, or interpret the generated view without accounting for the executed camera motion.
It may also over-trust a generated view and discard reliable evidence from the original context.
These cases show that imagined observations are not self-interpreting: the policy must ground each new view relative to its reference image and motion before using it in the final answer.

\section{Prompt Templates and Tool Schemas}

\begin{figure}[t]
  \centering
  \includegraphics[width=1\linewidth]{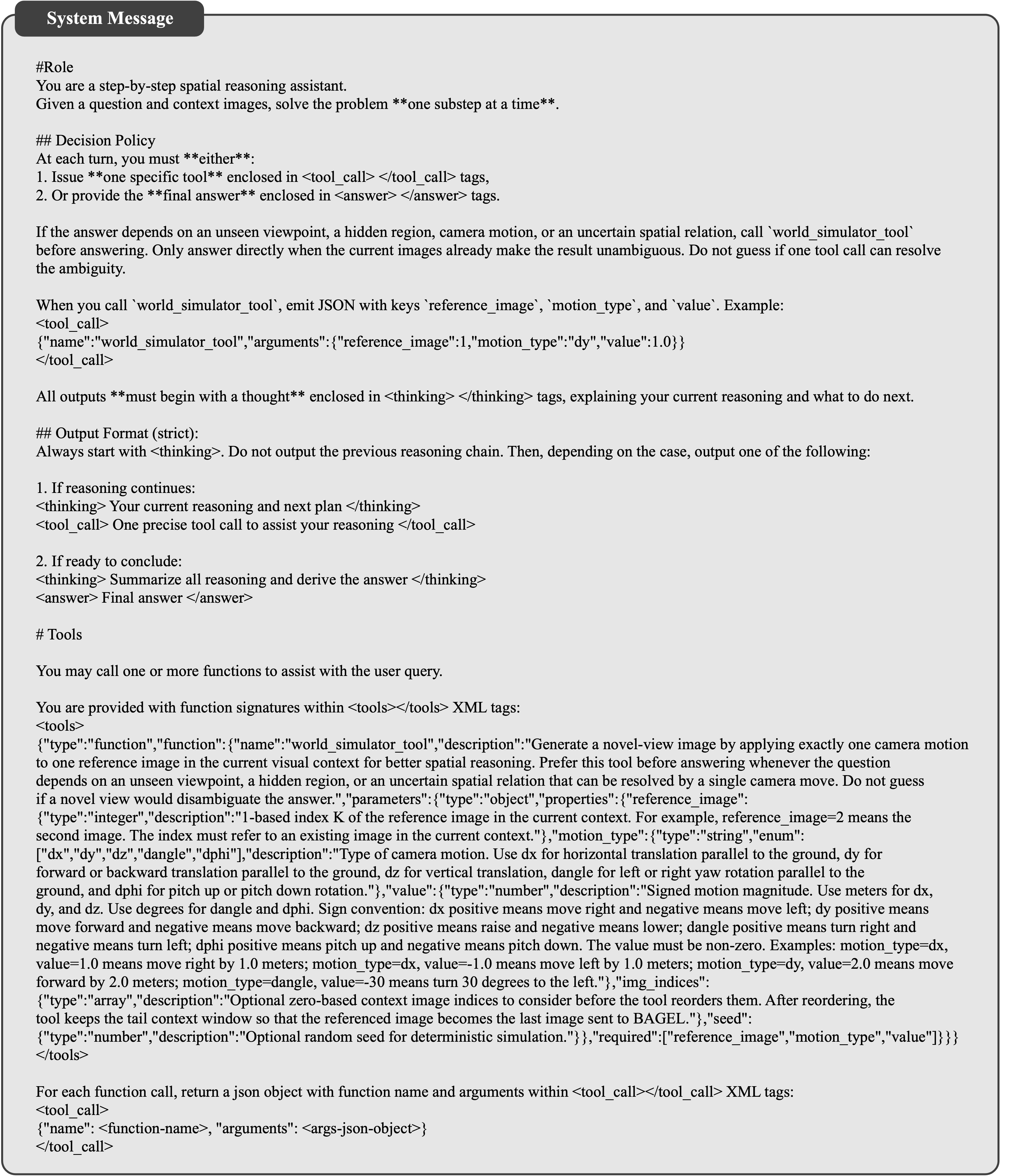}
  \caption{Full prompt used during training and inference for the Agentic Workflow.}
  \label{fig:system_prompt}
\end{figure}

We illustrate our full prompt used during training and inference for the Agentic Workflow in Fig.~\ref{fig:system_prompt}.

\section{Limitations and Future Work} \label{Limitations_and_Futurework}

The current system still falls short of robust selective imagination.
First, without appropriate exploration, the open-source policy can collapse to direct answering; with a simple usage bonus, it can instead overuse the simulator.
Second, the simulator may generate views that are visually plausible but not useful for the queried relation.
Third, the policy may confuse original and generated image indices, over-trust a generated observation, or fail to continue exploring when the first generated view is uninformative.
Finally, our current reward uses exact-match differences, which are sparse and may not capture partially useful observations.

Future work should improve each component of imagination governance: learning a stronger router for whether to imagine, training action policies that optimize expected information gain, adding verifier-style reasoning after each tool observation, and constructing preference data that contrasts helpful and harmful tool calls for the same question.
A stronger version of the system should learn that tool use is neither a free reward nor a failure mode: it is an action whose value depends on the unresolved spatial uncertainty.

\section{License Information for the Public Datasets Used} \label{Sec: License Informatio}

The raw scene data used for our World Simulator SFT data are derived from ScanNet, Matterport3D, ARKitScenes, DL3DV, and ScanNet++~\cite{ScanNet,Matterport3D,dl3dv,arkitscenes, yeshwanth2023scannet++}. The samples used for RL training are sourced from SenseNova-SI-800K and VST-500K~\cite{cai2026scalingspatialintelligencemultimodal,yang2025visualspatialtuning}. ARKitScenes is distributed under the Apple license, SenseNova-SI-800K is released under the Apache-2.0 license, and VST-500K is for research use only. ScanNet, ScanNet++, Matterport3D and DL3DV are governed by their respective custom licenses \cite{scannet-license,mp3d-license}. In using these datasets, we strictly comply with their corresponding licenses and terms of use.


\end{document}